\documentclass[11pt, a4paper, twocolumn, gdm]{google}
\usepackage{times}
\usepackage{latexsym}
\usepackage{amsmath}
\usepackage{graphicx}
\usepackage{booktabs}
\usepackage{multirow}
\usepackage{amssymb}
\usepackage{natbib}
\setcitestyle{authoryear,round}
\uselogo{} 
\usepackage{kotex} 
\usepackage{xcolor}

\title{Are you speaking my languages? \\
On spoken language adherence in multimodal LLMs}

\author{Hyungwon Kim}
\author{Kandarp Joshi}
\author{Lillian Zhou}
\author{Pavel Golik}
\author{Petar Aleksic}
\affil{Google DeepMind}
\affil{\texttt{\{clintk, kandarpj, lqz, golik, apetar\}@google.com}}

\date{}

\begin{document}
%
\begin{abstract}
While Large Language Model (LLM) based Automatic Speech Recognition (ASR) enables seamless multilingual use, models often misidentify the output language, compromising transcription fidelity and downstream application quality. To preserve flexibility and code-switching capabilities, we propose a soft prompting approach that hints at potential spoken languages without strictly constraining the output. We formally define this challenge as a lack of language adherence, introduce a novel metric to quantify violations, and evaluate three mitigation strategies: (1) zero-shot prompting for robust guidance under uncertainty, (2) supervised fine-tuning (SFT) to improve prompt adherence, and (3) Chain-of-Thought (CoT) reasoning to enforce adherence during decoding. We present a comparative analysis of these methods across multiple languages, evaluating effectiveness in reducing the language violation while maintaining overall ASR performance. Finally, we discuss trade-offs to guide strategy selection under various compute constraints.
\end{abstract}

\maketitle
\section{Introduction}
\label{sec:intro}

\subsection{Background}
Large Language Models (LLMs) have significantly advanced multilingual Automatic Speech Recognition (ASR), enabling flexible, zero-shot transcription across many languages, including code-switching (where multiple languages are spoken in the same utterance). While flexible, such systems often misidentify the target language in short or noisy segments. Robust multilingual support requires the integration of user context; specifically, language hinting guides the model toward the expected language without restricting the user's ability to code-switch. This approach successfully balances transcription accuracy with linguistic freedom.

\subsection{Problem Statement}
Lack of language adherence fundamentally degrades transcription fidelity, introducing errors that distort the original meaning. This unreliability directly compromises downstream tasks dependent on accurate ASR, such as machine translation, sentiment analysis, and command systems. Furthermore, poor adherence creates a jarring negative user experience. More than typical ASR errors, erroneous foreign-language text can feel amateurish or be perceived as insensitive to cultural contexts or biased towards a user's accent, eroding trust in the technology. We need a way to mitigate these pitfalls while supporting flexible multilingual engagement.

\subsection{Proposed Approach and Contributions}
To systematically address the issue of poor language adherence, this paper introduces a multifaceted approach encompassing formalization, measurement, and mitigation. Our primary contributions are as follows.

First, we formally define ``Language Adherence Violation'' and propose a novel metric to quantify these occurrences, providing a standardized evaluation method. Then, we discuss the inherent difficulty of obtaining reliable language preferences.

Second, we propose and investigate three non-mutually exclusive mitigation strategies to improve language adherence while balancing flexibility and robustness to incorrect language signals:
\begin{itemize}
    \item Prompt Engineering: Using carefully designed prompts to guide the LLM's focus to a target language while assessing its robustness to imperfect signals.
    \item Supervised Fine-Tuning (SFT) with instruction: Employing language-adherence prompts during the SFT process to ingrain the desired transcription behavior.
    \item Chain-of-Thought (CoT): Implementing a reasoning step that compels the model to first identify and declare the spoken language before transcribing.
\end{itemize}

Finally, we conduct a comprehensive comparative analysis of these methods across monolingual and code-switching datasets, evaluating the trade-offs between language signal accuracy and model robustness. To maintain a focused comparison of prompting-based methods, we do not explore reinforcement learning approaches.

\section{Related Work}
\label{sec:related_works}

\subsection{Multilingual and Code-Switching ASR}
\label{subsec:langid}
Some early multilingual neural ASR models used separate output layers per language~\citep{scanzio2008}, requiring the target language to be known a priori. Others appended a one-hot language ID embedding to input features, allowing the model to learn language-specific biases~\citep{li2018}. While language ID can be estimated from audio~\citep{cole1989,ma2002,lopez2014,bazazo2023}, integration with streaming ASR is difficult due to the latency required for confident estimation. To address this, \citet{waters2019,zhang2022} utilized parallel language ID modules to provide running estimates to an RNN-T model, while more sophisticated solutions combine identification and verification to reconcile signals~\citep{kim2025}.

\citet{watanabe2017} proposed another approach to model acoustic language ID and ASR jointly: they modify the training data to include the language tag as the first word in the reference transcript, teaching the model to predict the language before outputting text. While this is close to how today's LLMs approach multitasking, the encoder was based on a bi-directional LSTM, which has access to the entire input audio before outputting its language ID estimate and is therefore not streaming-friendly. Still, it's a remarkable study as it discusses the language adherence rates explicitly. 

In the context of code-switching in Indic languages, \citet{emond2018,datta2020} proposed to transliterate all data into a single common script, which gives rise to positive cross-lingual effects and helps with under-resourced languages. While this works for modeling certain language pairs, restoring the original spelling for rendering of the ASR hypothesis might be infeasible in the most general case.

\subsection{Language Identification in Speech Processing}
\label{sec:textlangid}
To monitor unconstrained multilingual ASR or LLM outputs, robust text language identification (LangID) is essential, often utilizing character-level N-gram classifiers~\citep{cavnar1994,caswell2020}. While effective for long sentences, these classifiers struggle with short inputs—such as "ja" (a word that exists in over ten languages), and require additional context like user profiles or conversation history for accuracy. Furthermore, it needs to be flexible enough to allow for benign use of proper names (e.g.~``Apple'' in a French sentence about the company) or loan words (e.g.~``Download'' in a German sentence about mobile apps) and to identify individual character spans in a code-switching environment.

While only a few languages have unique writing systems (e.g. Armenian, Georgian, Greek), many languages share large portions of their alphabets for historical reasons. For example, 22 out of 24 official languages of the European Union use Latin script. This increases mutual intelligibility: even if a speaker of French doesn't understand some Italian word, at least they recognize most characters and might even be able to guess pronunciation and meaning based on context. Conversely, a French speaker who is exposed to Arabic or Chinese script is less likely to do so. This simple observation suggests that it can be more practical to focus on identifying unexpected \textit{script} rather than \textit{language}. This can be done very efficiently based on Unicode ranges~\citep{qasim2024}.

\subsection{Large Language Models in ASR}
Following early modality fusion approaches \citep{sun2019,zheng2021,bapna2022,wang2023}, most of today's leading LLMs are natively multimodal. They are trained on text, audio and images, tokenized with modality-specific encoders to produce features in a unified space or tokens from a single flat vocabulary. In order to use an LLM for speech recognition, it is usually prompted with ``Transcribe the following audio:'', followed by the audio tokens. While it's possible to process some tasks in the same LLM call (e.g. take speech as audio input and immediately generate audio tokens as response), there are still applications that choose to chain different models.
In a cascaded system like this, where the ASR hypothesis is used as input to a separate model (e.g. fed into a different LLM or a machine translation system), the quality of the downstream application can be greatly harmed by language adherence issues in the ASR transcription. Additional context is required to disambiguate between the music genre ``Soul'' and the capital of South Korea ``서울'' (in English, ``Seoul'').

\subsection{Explicit Language Control in Generative Models}
While the controlled generation of formal languages such as code has been studied extensively (e.g. with the help of finite state automata~\citep{koo2024}), no established solution exists for natural languages yet. Since decoding LLMs with beam search is often considered expensive, hybrid solutions seek to modify the logit or probability values heuristically in order to support or suppress subsets of the vocabulary, before the decoder samples the next token~\citep{dathathri2019}.

\section{Measuring Language Adherence}
\label{sec:measurement}

\subsection{Language Adherence Metric}

For offline evaluation purposes, we assume that each utterance in the test set can be uniquely annotated with a set of languages $L_{ref}$ that appear in the audio. Usually, the set contains only one or two languages.

As mentioned in Sec.~\ref{sec:textlangid}, an external text LangID tool can be used to obtain the set of languages $L_{hyp}$ that appear in the generated output. Since the hypothesis can contain recognition errors, we don't require the sets to match perfectly. 

\textbf{Definition} The generated model output violates the language adherence whenever $L_{hyp} \not\subseteq L_{ref}$.

Thus, for a dataset with $N$ utterances, the \textit{Language Adherence Violation Rate} (LAVR) is
\begin{equation}
\textrm{LAVR} = \frac{1}{N}\sum_{i=1}^N \mathbb{I}(L_{hyp, i} \not\subseteq L_{ref, i})
\end{equation}
\label{metric:LAVR}
where $\mathbb{I}$ is an indicator function and $L_{hyp, i}$ and $L_{ref, i}$ are the set of languages of $i$-th utterance. 

In the context of this study, measuring language adherence violations (sometimes referred to as ``(local) language hallucinations''), we operationally define ``language'' by its corresponding canonical \textbf{script}; thus, a ``language violation'' occurs when the ASR system produces a script inconsistent with the target language's standard orthography. Thus the set of expected ``languages'' of a German test set is actually defined as a set of acceptable characters $L_{ref} := [\text{a-zäöüß}]$. Similarly, $L_{hyp}$ is the set of characters that appear in the generated output. For example, both ``Danke schön'' and ``Thank uoy'' are language adherent for German (note the spelling typo in ``uoy''), but ``Danke sçhön'' is not due to `ç'. In addition, we treat basic punctuations and digits in ASCII characters as ``neutral'' and exclude them from measurement, while language specific punctuations (e.g., ``¿'') are retained. Note that a single unexpected character in $L_{hyp}$ marks the entire utterance as a violation. This is an acceptable simplification, given the homogeneous length distribution of our evaluation sets. 

Our ultimate goal is to reduce the ratio of utterances violating the language adherence under uncertainty about user preferences without increasing the word error rate (WER). A low rate of language adherence violations suggests a high level of control over the behavior of a multilingual system.

\subsection{Metric Scope and Limitations}

As discussed in Sec.~\ref{sec:textlangid}, a language adherence metric based on word-level text lang ID can be ambiguous. We found that even the reference transcripts of many real-world test sets have a non-zero amount of language adherence violations. This is partly due to loan words (e.g. ``Kindergarten'') or proper names (e.g. ``Versace'').

On the other hand, as a coarse-grained metric, the rate of character-level language adherence violations in reference transcripts is very close to zero. The only violations are due to foreign words spelled in the original script, such as ``jalape\~{n}o'' or ``Beyonc\'{e}''. Thus many Finnish and English words will be considered ``acceptable'' in a German test set. This is consistent with the notion of surprise in user perception discussed in Sec.~\ref{sec:textlangid}. 

The character-level metric would miss a hypothetical bug that dropped all umlauts from German vowels. It would also remain zero if the decoded text consisted of nothing but the word ``hello'' or some gibberish like ``asdf''. This is a limitation of the proposed metric, and can be thought of as trading precision for recall: \textit{if} there is a character that would likely draw the user's attention, \textit{then} we want to make sure to flag it. It should always be complemented by the standard WER between reference and hypothesis.

\subsection{Online Metrics}
\label{sec:online_metrics}

Measuring performance on real-world traffic is challenging because manual transcription is expensive and scaling is difficult. 
We propose using user language settings and past conversations as a proxy for $L_{ref}$, though this approach has limitations. For instance, a bilingual user might speak French but receive a Spanish misrecognition that goes unflagged if both languages are in their settings; conversely, a user with English-only settings who naturally code-switches into Hindi will trigger a false violation. 
These scenarios, including new language learning and international travel, demonstrate the limits of this formalism. Consequently, we recommend monitoring relative changes in language adherence rather than striving for an absolute zero.




\section{Proposed methods and experimental results}
\label{sec:methods_and_results}

As discussed in Sec.~\ref{sec:measurement}, accurate identification of the spoken language is not always possible and the contextual signals can be misleading.
Recognizing this challenge, we explore three mechanisms to improve language adherence. 
While the methods can be implemented independently and combined in different ways, we choose to evaluate them sequentially, adopting the best outcome for the next stage.

\subsection{Experimental Setup}
\label{subsec:exp_setup}

\textbf{Evaluation Datasets} We collected two classes of datasets: mono-lingual and code-switching. The monolingual datasets comprise about a few thousands user queries per language, sourced from real-world interactions with a production-grade AI agent. Most utterances range in length from 5 to 20 words. For code-switching, we synthesized audio from a dataset of about 10,000 anonymized queries where users code-switch between English and another language with diverse voices. These utterances typically range from 5 to 10 words in length. See the Appendix for statistics and sample utterances of the evaluation dataset.


We report results for English, French, Hindi, Korean, and their pairing with English in the main text; findings for additional languages are provided in the Appendix. 

\textbf{Baseline Model} 
We use Gemini Flash lite 2.0 as the foundation model. Specifically, the baseline model for zero-shot experiments is a proprietary variant of the Gemini 2.0 Flash Lite, fine-tuned for general ASR tasks. It’s a deep transformer-based LLM trained on very large amounts of transcribed speech in all languages discussed in this work. For the SFT and CoT experiments, we further fine-tune it using a proprietary collection of transcribed speech data, consisting primarily of mono-lingual and code-switching single-utterance recordings.

\textbf{Evaluation Metric} For each proposed method, we compute character-level LAVR, defined in Eq.~\ref{metric:LAVR}, and word error rate (WER). We evaluate performance by varying the language hints in the prompt across four scenarios relevant to real-world use:
\begin{itemize}
    \item \textbf{no-hint}: No language hint is provided.
    \item \textbf{correct}: The prompt contains the correct language(s) only. This is an idealistic oracle condition.
    \item \textbf{distractor}: The prompt contains an incorrect language only (Japanese for Korean, Spanish for French and Hindi).
    \item \textbf{mix}: The prompt contains both the correct language and the distractor language.
\end{itemize}
For code-switching datasets, the \textbf{correct} case includes both spoken languages, while the \textbf{distractor} case includes only the distractor of the non-English language, and the \textbf{mix} case includes the non-English language and its distractor. For \textbf{no-hint} case, the prompt is always ``Transcribe the following speech segment: ''. As discussed in Sec. \ref{sec:online_metrics}, getting the correct language signal is not always possible and thus it is important that the system has to perform reasonably well under non-ideal scenarios.

Given the ubiquity of English proper names and abbreviations (e.g. ``Nvidia'', ``RAM'' etc.), we always consider Latin characters $[\text{a-z}]$ to be acceptable in the output when computing the language adherence metric. Therefore, ``Netflix 신작'' (Netflix new release) as a Korean output does \textit{not} violate language adherence.

\textbf{Model Selection} We select the best zero-shot prompt and SFT mixture based on the language adherence performance on dedicated short utterance datasets (1,500 English, 3,000 Korean utterances, each utterance is one word), since they are challenging due to phonetic ambiguity and lack of context. The best model is the one with the overall smallest LAVR across scenarios, and the smallest LAVR difference between the \textbf{correct} and \textbf{distractor} hint scenarios, indicating robustness.

\subsection{Method 1: zero-shot (ZS) Language Hint Prompting}
\label{subsec:zero-shot_method}
The first approach leverages in-context learning capabilities of LLMs via zero-shot prompting. We provide a language \textit{hint} to bias the model towards the target language, influencing its hypothesis space. An ideal prompt must balance guiding the model while allowing it to adhere to the audio evidence if the hint is incorrect.

We tested variants of three prompt styles:
\begin{itemize}
    \item \textbf{P1}: Transcribe the following speech segment in \texttt{<languages>}:
    \item \textbf{P2}: The following speech segment is spoken by someone who knows \texttt{<languages>}. Transcribe the following speech segment:
    \item \textbf{P3}: Transcribe this speech segment. It may contain a mix of \texttt{<languages>} and other languages.
\end{itemize}

The LAVR results on our short-utterance datasets are shown in Table \ref{tab:zero-shot_result}. Based on these results, we selected P3 for our main experiments, because it proved the most robust against incorrect hints. Phrasing variations within each prompt style had a negligible impact on the results.

\begin{table}[h!]
\centering
\begin{tabular}{lcccccc}
\toprule
& \multicolumn{3}{c}{English} & \multicolumn{3}{c}{Korean} \\
\cmidrule(lr){2-4} \cmidrule(lr){5-7}
Prompt & cor & dist & mix  & cor & dist & mix\\
\midrule
P1 & 1.7 & 22.5 & 5.9 & 0.0 & 6.5 & 7.3  \\
P2 & 1.9 & 3.0 & 1.8 & 0.0 & 6.2 & 0.1\\
P3 & 2.3 & 2.0 & 1.8 & 0.0 & 3.3 & 0.0\\
\bottomrule
\end{tabular}%
\caption{LAVR [\%] (lower is better) of short utterance datasets by prompts.  ``cor'' for correct and ``dist'' for distractor.}
\label{tab:zero-shot_result}
\end{table}

\subsection{Method 2: Supervised Fine-Tuning (SFT)}
\label{subsec:SFT}
To explicitly teach the model to follow a language-hinted prompt, we apply the SFT method to minimize the cross entropy loss for token prediction. The primary objective is to ensure adherence to language hints while preserving robustness against potentially erroneous cues. Note that the SFT model has the same latency as the baseline model. 

\textbf{Prompting strategy for training}
For a given training set, we include a system instruction using the \textbf{P3} prompt from Sec. \ref{subsec:zero-shot_method}.
To ensure robustness of the trained model, we randomly vary the hinted language using the four categories defined in Sec.~\ref{subsec:exp_setup}. For \textbf{distractor} and \textbf{mix} cases, we add up to three (60\% one, 30\% two, and 10\% three) randomly selected languages from a pool of 56 languages.

\textbf{Mixture of training prompts}
We experimented with several training data configurations, varying the distribution of hinted language types. Applying the same model selection criteria (see the Appendix for exact results), we chose the model trained on a mixture of 10\% no-hint, 40\% of correct, 35\% of distractor only, and 15\% mixed prompts. 

\subsection{Method 3: Chain-of-Thought (CoT) Prompting}
\label{sec:CoT}
Inspired by~\citet{watanabe2017}, our final method uses a Chain-of-Thought (CoT) approach to force the model to explicitly reason about the language before transcribing. For each example, we appended the following specific CoT prompt, including special control tokens to separate the language identification step from the transcription:

\begin{quote}
  Think about the languages of the
  speech and transcribe it in those languages.
\end{quote}

The rationale is to narrow the sampling space for the transcription task by first committing to a language. Formally, the generated word sequence is conditioned on the (estimated) language IDs. While this technique can be considered a variant of SFT, its distinct reasoning process merits separate analysis. Because the model emits only a few additional tokens (the language name), the impact on average decoding time is negligible. It should be noted that our models operate in a non-streaming fashion, where the full audio context is available prior to decoding. Conversely, a streaming implementation might face increased latency if the model defers output to stabilize its language prediction.


For constructing the CoT training data, we prepended the correct language to the reference transcript, enclosed in control tokens. The mixture of prompts is comprised of 90\% \textbf{distractor-only} and 10\% \textbf{no-hint}. This composition ensures the model learns the correct language signal, while all other training parameters remain consistent with the SFT experiments.


\begin{table}[t!]
\centering
\small 
\setlength{\tabcolsep}{4pt} 
 \resizebox{\columnwidth}{!}{%
\begin{tabular}{llcccc}
\toprule
& & \multicolumn{4}{c}{\textbf{Monolingual}} \\
\cmidrule(lr){3-6}
\textbf{\rotatebox{90}{Lang}} & \textbf{Exp} & \textbf{correct} & \textbf{distractor} & \textbf{mix} & \textbf{no-hint} \\
\midrule
\multirow{3}{*}{\rotatebox{90}{English}} & ZS & 0.8 (6.8) & 0.7 (7.2) & 0.7 (7.7) & 1.0 (6.9) \\
                         & SFT                & 0.9 (6.5) & 0.9 (6.3) & 1.0 (6.5) & 1.2 (7.7) \\
                         & CoT                & 1.2 (7.6) & 1.1 (6.8) & 1.1 (6.6) & 1.8 (7.5) \\
\midrule
\multirow{3}{*}{\rotatebox{90}{French}} & ZS  & 0.2 (9.6) & 1.2 (12.4) & 0.3 (11.0) & 2.2 (10.6) \\
                        & SFT                 & 0.9 (11.2) & 1.6 (12.0) & 0.9 (11.3) & 2.6 (11.1)\\
                        & CoT                 & 1.1 (10.6) & 1.5 (11.0) & 1.2 (10.6) & 2.6 (16.7) \\
\midrule
\multirow{3}{*}{\rotatebox{90}{Hindi}} & ZS   & 0.0 (12.2) & 1.1 (13.0) & 0.0 (11.2) & 0.6 (11.4) \\
                       & SFT                  & 0.0 (11.2) & 0.5 (11.5) & 0.0 (11.3) & 0.4 (11.4) \\
                       & CoT                  & 0.0 (11.2) & 0.1 (11.5) & 0.0 (11.1) & 0.3 (29.1) \\
\midrule
\multirow{3}{*}{\rotatebox{90}{Korean}} & ZS  & 0.4 (11.0) & 3.5 (11.7) & 0.6 (11.0) & 1.7 (11.3) \\
                        & SFT                 & 0.9 (11.3) & 1.7 (11.4) & 1.0 (11.3) & 1.8 (10.9) \\
                        & CoT                 & 0.6 (11.6) & 1.2 (11.2) & 0.8 (11.8) & 2.2 (14.5) \\
\bottomrule
\end{tabular}%
}
\caption{Monolingual Results. LAVR (\%) and WER (CER for Korean, \% in parenthesis) by language, model, and language hint type.}
\label{tab:mono_results}
\end{table}

\begin{table}[t!]
\centering
\small 
\setlength{\tabcolsep}{4pt} 
 \resizebox{\columnwidth}{!}{%
\begin{tabular}{llcccc}
\toprule
& & \multicolumn{4}{c}{\textbf{Code-switching (with English)}} \\
\cmidrule(lr){3-6}
\textbf{\rotatebox{90}{Lang}} & \textbf{Exp} & \textbf{correct} & \textbf{distractor} & \textbf{mix} & \textbf{no-hint} \\
\midrule

\multirow{3}{*}{\rotatebox{90}{French}} & ZS & 0.1 (31.1) & 0.2 (31.3) & 0.1 (30.8) & 0.1 (31.9) \\
                        & SFT       & 0.1 (30.4) & 0.1 (29.7) & 0.1 (29.7) & 0.2 (31.2) \\
                        & CoT       & 0.1 (28.9) & 0.1 (29.0) & 0.1 (29.0) & 0.2 (33.9) \\
\midrule
\multirow{3}{*}{\rotatebox{90}{Hindi}} & ZS & 0.0 (24.4) & 0.4 (25.1) & 0.0 (24.3) & 0.4 (25.9) \\
                       & SFT       & 0.1 (25.2) & 0.4 (26.2) & 0.1 (25.1) & 0.8 (26.9) \\
                       & CoT       & 0.1 (26.0) & 0.3 (25.9) & 0.1 (25.7) & 0.4 (37.3) \\
\midrule
\multirow{3}{*}{\rotatebox{90}{Korean}} & ZS & 0.1 (19.2) & 6.1 (21.9) & 0.6 (19.2) & 0.7 (20.7) \\
                        & SFT       & 0.4 (19.7) & 2.2 (20.8) & 0.8 (20.3) & 1.0 (21.3) \\
                        & CoT       & 0.4 (19.0) & 0.9 (19.5) & 0.5 (18.9) & 0.7 (21.0) \\
\bottomrule
\end{tabular}%
}
\caption{Code-switching (with English) Results. LAVR (\%) and WER (CER for Korean, \% in parenthesis) by language, model, and language hint type.}
\label{tab:cs_results}
\end{table}

\subsection{Comparative Results}
The final LAVR and WER for the three methods are summarized in Tables~\ref{tab:mono_results} and~\ref{tab:cs_results}. As expected, a consistent observation across nearly all evaluated scenarios is the superior performance associated with the ``correct'' language hint as this consistently produced the most favorable outcomes in both LAVR and WER metrics for all three methods.
Furthermore, a clear trend emerged: providing at least one correct language hint (i.e., the ``correct'' and ``mix'' conditions) significantly outperforms the ``no-hint'' and ``distractor'' conditions across both metrics. This pattern underscores the critical importance of obtaining a reliable estimation of the spoken language (e.g.~by model prediction and/or other user metadata).

On the whole, the three tested approaches demonstrated broadly comparable performance levels when supplied with identical language hinting prompts. A notable exception to this trend was the elevated WER observed for both the SFT and CoT methods under the ``no-hint'' prompt condition. This performance degradation is likely attributable to catastrophic forgetting. The baseline model was previously fine-tuned extensively using the ``no-hint'' prompt; consequently, the reduced proportion (10\%) of ``no-hint'' data utilized during the SFT and CoT training phases appears to have degraded performance for this specific ``no-hint'' scenario. These results strongly suggest that advancements in prompt exploration and, most critically, the accurate inclusion of the spoken language in the prompt, are the most impactful factors for optimizing system performance.




\section{Conclusions}
We introduce a novel quantitative metric, the Language Adherence Violation Rate (LAVR), to systematically evaluate language adherence in multilingual ASR systems. Recognizing the inherent challenge of obtaining an accurate language signal, we investigated three distinct strategies, zero-shot prompting, SFT, and CoT, to enhance system robustness against potentially erroneous language hints. Our findings, based on both LAVR and WER metrics, demonstrate that the zero-shot prompting strategy is as effective as the SFT and CoT approaches, reinforcing the critical importance of a correct language hint.

We also observed that when a correct language hint is present, the inclusion of a distractor language (``mix'' condition) has a negligible negative impact and still significantly outperforms the ``no-hint'' scenario. Conversely, providing only a distractor language hint yields performance that is generally inferior to the ``no-hint'' condition. Therefore, these results strongly advocate for the implementation of upstream mechanisms, such as dedicated Language ID models or metadata analysis, to predict at least one spoken language with high confidence.


\section{Acknowledgments}
We thank Diamantino Caseiro for encouraging us to write this paper.

\bibliographystyle{abbrvnat}
\bibliography{refs_long_form}

\clearpage
\appendix
\section{Appendices}

\subsection{More information about the evaluation set}
We present the overall statistics and sample utterances for the mono-lingual and code-switching evaluation set used in the main text in Table ~\ref{tab:dataset-samples}. Transcripts are presented in the native script, except Hindi, where we transliterated to English from the original Hindi script in the dataset for presentation. In all evaluation, we used the native scripts.

To develop our code-switching dataset, we utilized a proprietary voice generator featuring ten diverse male and female voices per primary (non-English) language, randomly selecting one per utterance for each language pair. The voices were configured to maintain the accent of the non-English language; for instance, in the Korean-English dataset, English phrases are delivered with a Korean accent while the Korean phrases maintain native fluency.

\begin{table*}[ht]
\centering
\small
\begin{tabular}{lrr p{8.5cm}}
\toprule
\textbf{Language} & \textbf{Num Utterances} & \textbf{Total Hours} & \textbf{Sample Transcripts (Native Script)} \\ \midrule
English & 1,760 & 5.7 & ``Can you help me with some things that I can tell my grandchildren that they can say if they get teased at school?'' \vspace{0.5em} \newline
                             ``Do you know where tiger high school is?'' \vspace{0.5em} \newline
                             ``Tell me a kid's joke about robots.'' \\ \midrule
French & 3,152 & 2.1 & ``Je me suis fait griffer par mon chat à la main et c'est profond. Qu'est-ce que je dois faire?'' \vspace{0.5em} \newline
                            ``Des fois, je fais des choses que je regrette après.'' \vspace{0.5em} \newline
                            ``Peux-tu me...'' \\ \midrule
Hindi & 2,784 & 2.4 & ``ne ah instagram pe isaka phul post diya apane?'' \vspace{0.5em} \newline
                       ``dee dvisht telar ke naam ke ladake kee teen lain kee kahaanee batao'' \vspace{0.5em} \newline
                         ``tumase milakar achchha laga.'' \\ \midrule
Korean & 6,448 & 4.5 & ``학생들도 시내 나가서 놀겠지?'' \vspace{0.5em} \newline
                             ``추천해 줄 거 있어?'' \vspace{0.5em} \newline
                             ``안산과 북악산을 대중교통으로 가는 방법 알려 주세요.'' \\ \midrule
French-English & 10,944 & 6.7 & ``c'est quoi will keep you posted en français'' \vspace{0.5em} \newline
                                ``do you know veste chauffante''\vspace{0.5em} \newline
                                ``Comment rendre apple watch écofriendly''\\ \midrule
Hindi-English & 10,160 & 7.2 & ``Mulethi powder face pr kya kya fayda dikhta hai'' \vspace{0.5em} \newline
                                ``Hair growth ya hairfall kam ke liye best kya h''\vspace{0.5em} \newline
                                ``Aur bhai bata everything is ok?''\\ \midrule
Korean-English & 9,968 & 7.9 & ``Pot roast는 어떤 요리야?'' \vspace{0.5em} \newline
                                ``주피터 노트북에서 import numpy 쓰러면 어떻게 해야하지?''\vspace{0.5em} \newline
                                ``stakeholder implications가 뭐야?''
                             \\ \bottomrule
\end{tabular}
\caption{Statistics and representative samples for the multilingual evaluation dataset. Hindi transcripts are transliterated to English from the original Hindi script in the dataset.}
\label{tab:dataset-samples}
\end{table*}

\subsection{Model selection for SFT mixture}
\label{appendix:sft_mixture}
In this section, we report the performance on short utterance dataset when varying the SFT training mixture, defined by the ratio of correct, distractor, mix, and no-hint prompt conditions. We evaluated the following candidates.
\begin{table}[h!]
\centering
\resizebox{\columnwidth}{!}{%
\begin{tabular}{lcccc}
\toprule
Mixture & correct & distractor  & mix & no-hint \\
\midrule
M1 & 0.4 & 0.25 & 0.25 & 0.1  \\
M2 & 0.4 & 0.35 & 0.15 & 0.1  \\
M3 & 0.4 & 0.3 & 0.2 & 0.1 \\
M4 & 0.4 & 0.3 & 0.1 & 0.2 \\
M5 & 0.4 & 0.2 & 0.2 & 0.2 \\
M6 & 0.4 & 0.1 & 0.3 & 0.2 \\
\bottomrule
\end{tabular}%
}
\caption{SFT training mixture's prompt type ratio.}
\end{table}

We selected prompt P3 from Sec.~\ref{subsec:zero-shot_method} for training. The performance on short English and Korean utterances is given in Tables~\ref{tab:a1_en} and~\ref{tab:a1_ko}.

\begin{table}[h!]
\centering
\resizebox{\columnwidth}{!}{%
\begin{tabular}{lcccc}
\toprule
Mixture & correct & distractor  & mix & no-hint \\
\midrule
M1 & 3.6 & 3.1 & 3.3 & 5.6  \\
M2 & 3.0 & 3.0 & 2.4 & 5.4  \\
M3 & 3.3 & 23.6 & 18.5 & 8.9 \\
M4 & 2.2 & 4.8 & 2.4 & 6.0 \\
M5 & 3.5 & 14.7 & 19.8 & 16.9 \\
M6 & 3.4 & 24.3 & 19.8 & 16.9 \\
\bottomrule
\end{tabular}%
}
\caption{LAVR on short English data by SFT mixture.}
\label{tab:a1_en}
\end{table}

\begin{table}[h!]
\centering
\resizebox{\columnwidth}{!}{%
\begin{tabular}{lcccc}
\toprule
Mixture & correct & distractor  & mix & no-hint \\
\midrule
M1 & 0.0 & 1.3 & 0.1 & 1.1  \\
M2 & 0.1 & 1.0 & 0.2 & 1.3  \\
M3 & 0.1 & 0.8 & 0.1 & 1.1 \\
M4 & 0.0 & 2.0 & 0.1 & 0.7 \\
M5 & 0.1 & 2.2 & 0.1 & 1.6 \\
M6 & 0.0 & 1.9 & 0.0 & 0.7 \\
\bottomrule
\end{tabular}%
}
\caption{LAVR on short Korean data by SFT mixture.}
\label{tab:a1_ko}
\end{table}

Applying the model selection criteria discussed in Sec.~\ref{subsec:exp_setup}, we choose M2 as the final candidate.

\subsection{SFT with different base prompts.}
In the main text, the SFT training was done with the best prompt determined by the zero-shot experiments (P3 from Sec.~\ref{subsec:zero-shot_method}). Here we present how other prompts affect the performance of the SFT model on the short utterances datasets. The  mixture of language hints in training is held constant (M2 from Sec.~\ref{appendix:sft_mixture}).

\begin{table}[h!]
\centering
\resizebox{\columnwidth}{!}{%
\begin{tabular}{lcccc}
\toprule
Prompt & correct & distractor  & mix & no-hint \\
\midrule
P1 & 5.4 & 21.5 & 23.9 & 6.7  \\
P2 & 3.0 & 23.8 & 19.3 & 4.1  \\
P3 & 3.0 & 3.0 & 2.4 & 5.4 \\
\bottomrule
\end{tabular}%
}
\caption{LAVR on short English data by SFT prompt.}
\end{table}

\begin{table}[h!]
\centering
\resizebox{\columnwidth}{!}{%
\begin{tabular}{lcccc}
\toprule
Prompt & correct & distractor  & mix & no-hint \\
\midrule
P1 & 0.0 & 2.7 & 1.3 & 1.3  \\
P2 & 0.0 & 2.5 & 0.1 & 1.0  \\
P3 & 0.1 & 1.0 & 0.2 & 1.3 \\
\bottomrule
\end{tabular}%
}
\caption{LAVR on short Korean data by SFT prompt.}
\end{table}

We conclude that P3 performs best among the prompts tested with SFT. Although other prompts might improve with different mixtures (other than M2), we did not test all combinations. This added complexity highlights the cost-effectiveness of the zero-shot approach, as SFT requires tuning more variables, such as mixture composition.

\subsection{Performance on more languages}
In this section we evaluate the performance of the three proposed methods (ZS, SFT, and CoT) on more languages: German, Japanese and (Brazilian) Portuguese. For code-switching, all of them are mixed with English. The design of both monolingual and code-switching datasets is identical to the datasets used in the main results. Distractor languages are Dutch, Korean, and Spanish for German, Japanese, and Portuguese, respectively. The monolingual results are shown in Table~\ref{tab:mono_extra_lang} and code-switching results in Table~\ref{tab:cs_extra_lang}.

\begin{table}[t!]
\centering
\small 
\setlength{\tabcolsep}{4pt} 
 \resizebox{\columnwidth}{!}{%
\renewcommand{\arraystretch}{1.4}
\begin{tabular}{llcccc}
\toprule
& & \multicolumn{4}{c}{\textbf{Monolingual}} \\
\cmidrule(lr){3-6}
\textbf{\rotatebox{90}{Lang}} & \textbf{Exp} & \textbf{correct} & \textbf{distractor} & \textbf{mix} & \textbf{no-hint} \\
\midrule
\multirow{3}{*}{\rotatebox{90}{German}} & ZS & 0.2 (8.3) & 0.4 (9.1) & 0.2 (8.4) & 2.1 (9.0) \\
                         & SFT                & 1.4 (8.5) & 1.3 (8.8) & 1.2 (8.6) & 2.9 (8.8) \\
                         & CoT                & 1.1 (8.9) & 1.2 (9.3) & 0.9 (8.9) & 2.5 (16.5) \\
\midrule
\multirow{3}{*}{\rotatebox{90}{Japanese}} & ZS  & 1.7 (12.6) & 40.8 (35.5) & 14.0 (16.7) & 6.7 (13.4) \\
                        & SFT                 & 6.0 (12.8) & 14.2 (16.1) & 8.8 (13.5) & 6.1 (13.1)\\
                        & CoT                 & 4.6 (15.5) & 15.5 (18.3) & 8.5 (16.5) & 6.1 (17.5) \\
\midrule
\multirow{3}{*}{\rotatebox{90}{Portuguese}} & ZS   & 0.1 (6.8) & 0.2 (8.4) & 0.0 (6.9) & 1.0 (7.3) \\
                       & SFT                  & 0.6 (7.1) & 0.8 (7.7) & 0.4 (7.9) & 1.8 (7.6) \\
                       & CoT                  & 0.4 (7.3) & 0.6 (7.8) & 0.4 (7.4) & 1.3 (15.2) \\
\bottomrule
\end{tabular}%
}
\caption{Monolingual Results. LAVR (\%) and WER (CER for Japanese, \% in parenthesis) by language, model, and language hint type.}
\label{tab:mono_extra_lang}
\end{table}

\begin{table}[t!]
\centering
\small 
\setlength{\tabcolsep}{4pt} 
\renewcommand{\arraystretch}{1.4}
 \resizebox{\columnwidth}{!}{%
\begin{tabular}{llcccc}
\toprule
& & \multicolumn{4}{c}{\textbf{Code-switching (with English)}} \\
\cmidrule(lr){3-6}
\textbf{\rotatebox{90}{Lang}} & \textbf{Exp} & \textbf{correct} & \textbf{distractor} & \textbf{mix} & \textbf{no-hint} \\
\midrule

\multirow{3}{*}{\rotatebox{90}{German}} & ZS & 0.3 (19.5) & 0.4 (20.1) & 0.3 (19.5) & 0.5 (20.1) \\
                        & SFT       & 0.3 (19.4) & 0.3 (19.6) & 0.3 (19.4) & 0.6 (20.3) \\
                        & CoT       & 	0.4 (19.2) & 0.4 (19.5) & 0.4 (19.3) & 0.5 (22.4) \\
\midrule
\multirow{3}{*}{\rotatebox{90}{Japanese}} & ZS & 0.7 (16.2) & 18.2 (20.3) & 2.9 (16.5) & 1.6 (15.9) \\
                       & SFT       & 1.0 (15.9) & 3.9 (17.0) & 1.3 (16.0) & 1.6 (17.8) \\
                       & CoT       & 1.0 (15.1) & 2.8 (16.5) & 1.4 (15.2) & 1.2 (16.8) \\
\midrule
\multirow{3}{*}{\rotatebox{90}{Portuguese}} & ZS & 0.3 (16.4) & 0.3 (17.2) & 0.3 (16.4) & 0.7 (17.1) \\
                        & SFT       & 0.4 (16.9) & 0.6 (19.3) & 0.4 (17.1) & 0.9 (18.2) \\
                        & CoT       & 0.5 (16.6) & 0.5 (17.3) & 0.5 (16.1) & 0.9 (23.0) \\
\bottomrule
\end{tabular}%
}
\caption{Code-switching (with English) Results. LAVR (\%) and WER (CER for Japanese, \% in parenthesis) by language, model, and language hint type.}
\label{tab:cs_extra_lang}
\end{table}

We observe that the general trends established in the main experiments extend to these languages as well.

\end{document}